# Explainable Artificial Intelligence Techniques for Accurate Fault Detection and Diagnosis – A Review

Ahmed Maged[a,b,*], Salah Haridy[c,b], and Herman Shen[a]

[a] Mechanical Engineering Department, University of North Texas, Denton, Texas, USA

[b] Benha Faculty of Engineering, Benha University, Benha, Egypt

[c] Department of Industrial Engineering and Engineering Management, College of Engineering, University of Sharjah, Sharjah, United Arab Emirates

[*] Corresponding author: ahmed.maged@unt.edu

**Abstract**

As the manufacturing industry advances with sensor integration and automation, the opaque nature of deep learning models in machine learning poses a significant challenge for fault detection and diagnosis. And despite the related predictive insights Artificial Intelligence (AI) can deliver, advanced machine learning engines often remain a black box. This paper reviews the eXplainable AI (XAI) tools and techniques in this context. We explore various XAI methodologies, focusing on their role in making AI decision-making transparent, particularly in critical scenarios where humans are involved. We also discuss current limitations and potential future research that aims to balance explainability with model performance while improving trustworthiness in the context of AI applications for critical industrial use cases.

Keywords: Explainable Artificial Intelligence, Deep Learning, Machine Learning, Reliability

## 1. Introduction

The manufacturing industry has seen significant changes due to the increasing availability of sensors and their incorporation into existing systems. With the integration of advanced technologies and automation systems in manufacturing, available data formats are evolving to include complex data streams, such as sequences of images and videos, which can provide valuable information on the state of the machines and their components. The timely identification and diagnosis of faults can prevent equipment failure, reduce maintenance costs, improve system performance, and enhance safety. In recent years, Machine Learning (ML) algorithms, including Deep Learning (DL) models, have shown great promise in automating fault detection and diagnosis tasks. However, these models are often viewed as black boxes, making it challenging to understand how they arrived at their predictions. This lack of transparency can pose a significant barrier to adopting machine learning in safety-critical applications, where the interpretability and trustworthiness of the model are essential. The models may achieve high accuracy in many cases, but it is difficult to know whether they have learned relevant features. This lack of interpretability and transparency can be a significant obstacle to the adoption of machine learning in safety-critical applications, where understanding and trusting the model's predictions is essential [1].

To address this, the field of eXplainable Artificial Intelligence (XAI) has emerged, aiming to develop models with acceptable accuracy and interpretability. As shown in Figure 1, XAI techniques aim to provide insights into the inner workings of machine learning models, enabling users to understand and trust the model's predictions. In fault detection and diagnosis, XAI can help identify the root cause of a fault and help provide actionable recommendations for repair or maintenance. In recent years, the interest in XAI has seen a notable surge. Still, it is noteworthy that the term itself was coined by Van Lent et al. in 2004 [2], and the broader concept of explainability in ML has roots dating back to the 1970s, as referenced in Adadi and Berrada [3].

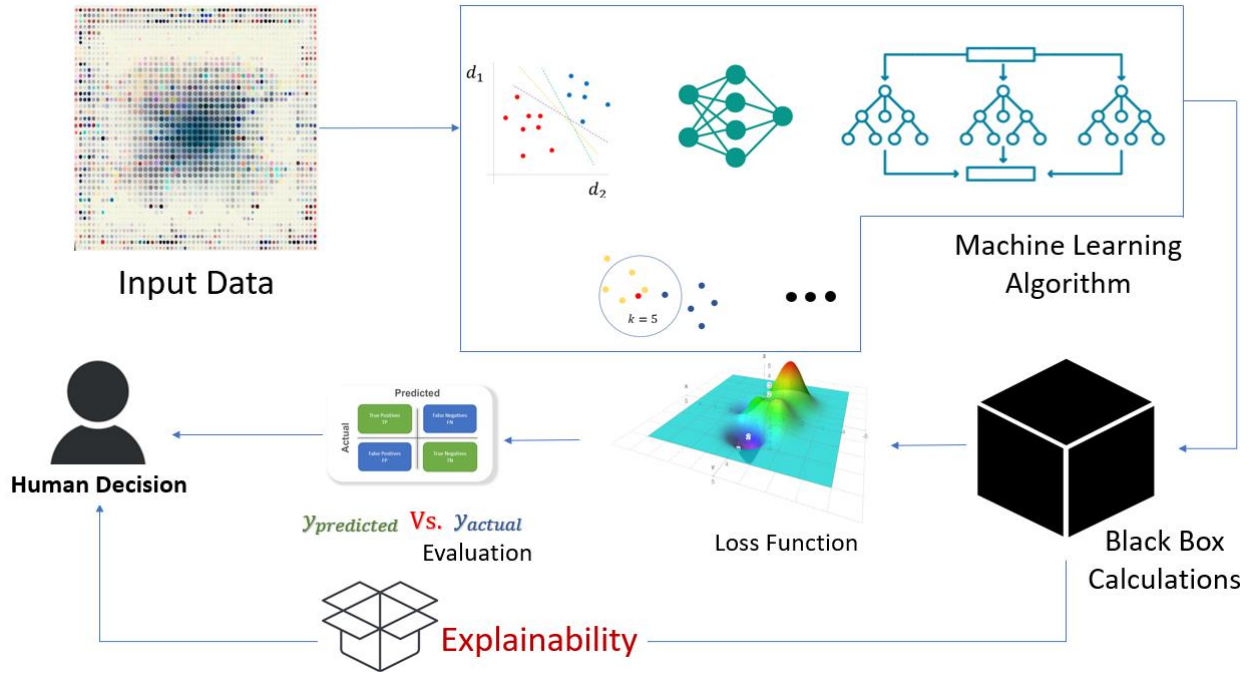

Figure 1. Schematic of XAI operational framework

While plenty of review papers explore various XAI methods, detailing their mechanisms and general applications, this paper is a bit more focalized. Our focus is on the application of XAI in fault detection and diagnosis, specifically within the manufacturing industry. This critical yet complex domain demands a targeted exploration of XAI's capabilities and limitations. Review papers such as [4] and [5] are recommended for readers interested in a broader understanding of XAI methods. We aim to bridge the gap in the literature by providing an in-depth analysis of how XAI can enhance fault detection and diagnosis, contributing to safer and more efficient manufacturing processes. Our research not only advances the theoretical understanding of XAI but also has practical implications for the manufacturing industry, contributing to safer and more efficient processes.

Hence, in this paper, we discuss some of the challenges and limitations of traditional machine learning models and present a comprehensive review of the state-of-the-art in XAI for fault detection and diagnosis in manufacturing and related fields. We examine the latest techniques for developing interpretable models. We also discuss the drawbacks of using XAI and the challenges of evaluating the explainability of machine learning models. Finally, we discuss the challenges and opportunities for future research and development in this exciting and rapidly evolving field.

In the next section, we give an overview of fault detection and diagnosis and a general introduction to XAI. Then, we review the literature on the different XAI methods and algorithms used in fault detection and diagnosis and explain how to choose between them.

## 2. Background

### 2.1 Fault Detection and Diagnosis

Fault detection and diagnosis methods have been crucial characteristics of safety-critical applications. Nonetheless, due to the demands of higher productivity and dependable operation, fault detection and diagnosis are incorporated into almost all sophisticated systems and pieces of equipment. A desirable fault detection and diagnosis system has the traits of monitoring overall system health as well as identifying and locating faults for the safe removal of faulty components within the system [6].

Most initial work on fault detection and diagnosis was carried out model-based approaches utilizing the adaptive observers and system identification models of the processes [7]. Model-based fault detection and diagnosis require an accurate mathematical model of the process and is the most suitable candidate for small systems with fewer inputs and an explicit mathematical model; however, its performance is drastically affected in the presence of unmodeled disturbances and uncertainties [8]. In contrast, data-driven approaches do not comply with the same rules. Accurate usage of the large amounts of collected data can help predict the occurrence of a fault or malfunction. These data-driven approaches usually involve statistical techniques such as control charts, signal processing, or machine learning. Machine learning has become increasingly popular due to its ability to handle large and complex datasets, its adaptability to changing conditions, and its ability to provide accurate and reliable predictions.

In general, ML-based fault detection can be achieved in three ways. The first involves identifying patterns of normal operation and distinguishing anomalies that indicate faults. The second involves training the model on both normal and non-normal data, enabling it to differentiate between the two patterns. The third method involves monitoring equipment output and predicting failures where a model is trained for regression, and the difference between predicted and actual values is monitored. These values, also called residuals, are compared to a certain threshold to determine if the equipment is faulty. For fault diagnosis, similar to the abovementioned second method, models are typically trained to distinguish between different faulty states in order to be able to predict them using input features.

Deep learning techniques are a type of artificial neural network that can learn complex patterns and relationships in data. They have been used in fault detection and diagnosis applications, particularly in image and speech processing applications.

Interested readers can refer to other review papers that discuss fault detection and diagnosis methods in detail, such as [7] and [9], since the focus of the paper is on XAI, rather than traditional ML methods used in fault detection and diagnosis.

## 2.2 Understanding Explainable Machine Learning

The primary goal of XAI in fault detection and diagnosis is to provide interpretable and transparent models that can help humans understand the decision-making process of the machine learning algorithm [10]. In fault detection, the goal is to identify when a fault has occurred, or when a component is likely to fail. This is typically done by analyzing sensor data from the machine and detecting any anomalies or deviations from normal behavior.

In that context, XAI can be helpful in two ways. First is the Model-Based Explanation, which occurs when a fault is detected by an algorithm and the goal is to interpret this detection. Understanding the reasons behind a specific detected fault is straightforward for inherently interpretable models (like logistic regression or shallow decision trees). For less interpretable models (such as complex neural networks), post-hoc techniques like Local Interpretable Model-Agnostic Explanations (LIME) and Shapley Additive Explanations (SHAP) are employed to clarify why the model identifies certain instances as faults. This ensures consistency between the model's detection and its explanation. Second is the Data-Centric Explanation, which is encountered when a fault is identified by an expert without a model, the focus is on understanding why the data is flagged as faulty, independent of any detection models. If an expert uses an unknown algorithm, there's a gap between the explanation and the detection method. This case emphasizes understanding faults in the data where no model is available, acting as surrogate methods for non-accessible fault detection models.

One might argue that XAI can only be used in the diagnosis phase rather than the detection phase. Nevertheless, it is important to recognize that the associations identified by machine learning algorithms do not inherently imply causality. There may be unobserved factors that are responsible for observed correlations among variables. In such cases, XAI can help understand the logic behind the predictions. For example, let us consider a CNC machine that operates within certain performance parameters that are monitored using multiple sensors, such as temperature, pressure, and vibration sensors. In the fault detection phase, the monitoring system may detect that the tool or the head is overheating and alert the user. The explanation provided by the XAI model in this case would focus on the features or variables that contributed to the fault detection, such as the temperature. In contrast, in fault diagnosis, we would use some diagnostic tool to determine the specific cause of the overheating problem, such as a malfunctioning thermostat, a coolant fluid, or a clogged nozzle. In this case, we may shut down the machine for maintenance or investigate further to determine the root cause of the temperature increase.

Moreover, the significance of XAI transcends technical details, encompassing broader aspects that influence its adoption and integration into such critical domains:

1. Users are more likely to trust a model when they can understand its workings and the rationale behind its decisions. This level of transparency is especially vital in high-stakes fields such as healthcare or finance, where the consequences of decisions can be significant.

2. XAI serves as a valuable tool for debugging, helping to identify and correct errors, biases, and other issues within the model. This not only enhances the system's overall performance and reliability but also ensures compliance with legal and regulatory requirements. For instance, regulations like the General Data Protection Regulation (GDPR) require organizations to disclose the logic behind automated decisions, thereby promoting accountability [10].

3. Ethical considerations emphasize the importance of XAI in promoting fairness and preventing discrimination. These considerations are crucial in addressing the broader ethical issues associated with artificial intelligence systems, ensuring that they operate in a manner that aligns with societal values and ethical norms.

**3. Techniques for XAI in Fault Detection and Diagnosis**

XAI provides interpretable and transparent explanations for the decisions made by machine learning models. It aims to bridge the gap between the complex inner workings of these models and the need for human comprehension and trust in their outputs. Machine learning interpretability methods can be classified based on various criteria as in Figure 2. A primary distinction differentiates whether interpretability is achieved through model design (intrinsic analysis) or post-hoc analysis.

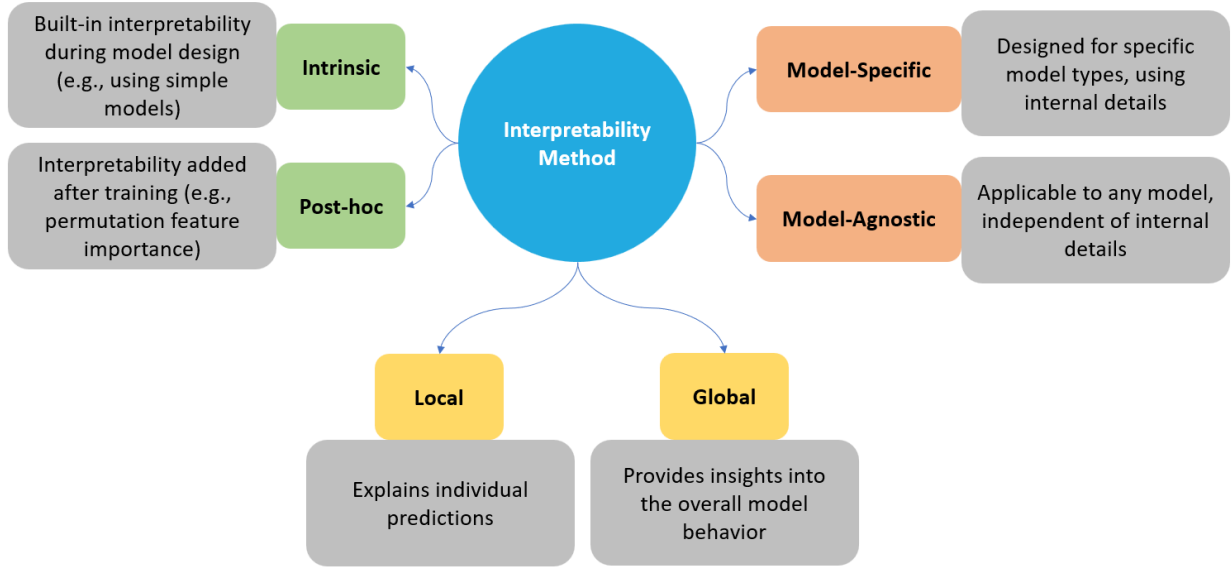

Figure 2. XAI different categories

Intrinsic methods embed interpretability into model construction by restricting complexity and utilizing intrinsically interpretable models like sparse linear models or short decision trees.

Interpretability arises from the model's transparent structure and inference process. Alternatively, post-hoc methods analyze a trained model to provide interpretation. Techniques like permutation feature importance are model-agnostic and can be applied after training. However, post-hoc approaches may also be used with intrinsically interpretable models, such as computing permutation importance for decision trees.

One can also categorize XAI methods as model-specific and model-agnostic. The distinction between model-specific and model-agnostic methods lies in their respective applicability to different types of machine learning models. Model-specific methods are specifically designed to work with a particular type or family of models, leveraging that model's the internal structure, characteristics, or properties to provide explanations. These methods rely on the unique assumptions, features, or computations of the specific model, allowing for tailored explanations. For example, interpreting the regression weights in a linear model is a classic example of model-specific interpretation. Alternatively, model-agnostic methods are designed to be independent of the underlying machine learning model. They can be applied to any type or family of models without requiring knowledge of the specific internal details. Model-agnostic methods aim to provide explanations by analyzing the model's input-output behavior without exploiting any model-specific characteristics. These methods are specifically developed to address the black-box nature of models and provide insights into their decision-making processes. Common model-agnostic techniques include LIME and SHAP since they can be applied to a wide range of models.

Finally, XAI can be distinguished based on whether the interpretation method elucidates the rationale behind an individual prediction (Local methods) or comprehensively elucidates the behavior of the entire model (Global methods).

The choice between intrinsic models and post-hoc explanation methods is not arbitrary but depends on several factors. These include the specific domain, the complexity of the problem, and, most importantly, the trade-offs between interpretability and model performance. In the field of fault detection, for instance, post-hoc explanation methods are generally more commonly used, indicating a preference for model performance over interpretability.

Intrinsic methods aim to build interpretability into the model design, as their structure and decision-making process are easy to understand. However, they may sometimes achieve a different level of accuracy than more complex models. Seeking to develop a hybrid approach that balances interpretability and accuracy, several researchers have proposed fault detection methods based on combining intrinsic models. Next, we detail the XAI techniques for fault detection and diagnosis.

For illustration purposes, we simulate a dataset containing 1000 samples with 10 features. Of these features, 5 are directly informative, meaning they correlate with either the faulty or non-faulty class, and 2 are redundant, derived as linear combinations of the informative features to reflect real-world correlations. The dataset represents a practical yet simplified representation of an actual fault detection system where machinery failures are common. Each fault occurrence is recorded with sensor data readings at the time of the fault. We later apply the XAI tools to it, whenever possible, in order to better grasp how they work. Figure 3. shows a scatter plot of the data where we can see that input features exhibit different relations across each other. We also present a linear plot and a histogram of the data in Figure 4, and 5, respectively. Notably, all features exhibit a high frequency of fluctuations without any visible trends or seasonal patterns. The histograms show that most features in the dataset display approximately Gaussian distributions while some of the variables are slightly skewed.

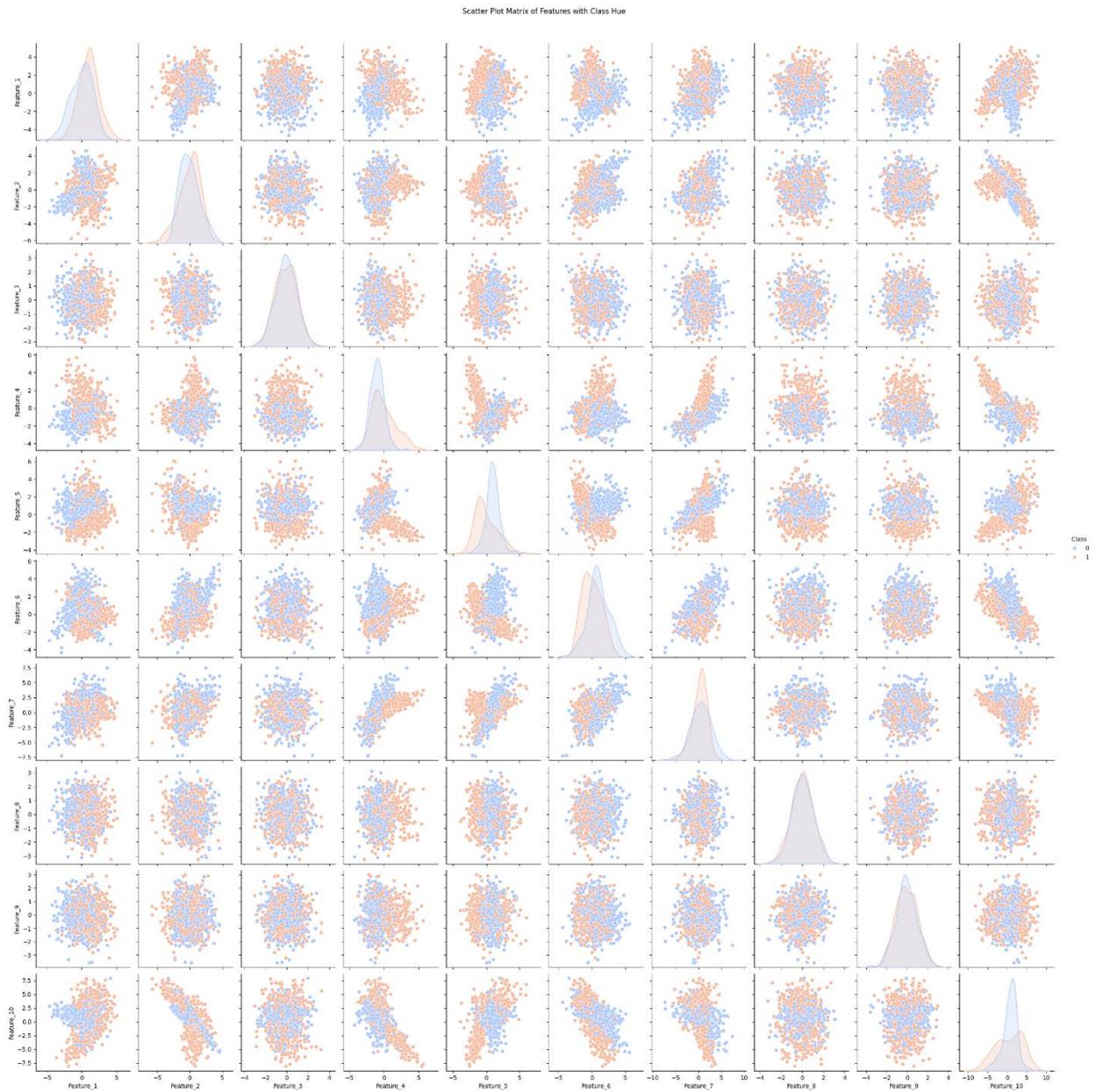

Figure 3. Scatter plot matrix of the simulated data with categories; faulty (orange) non-faulty (blue)

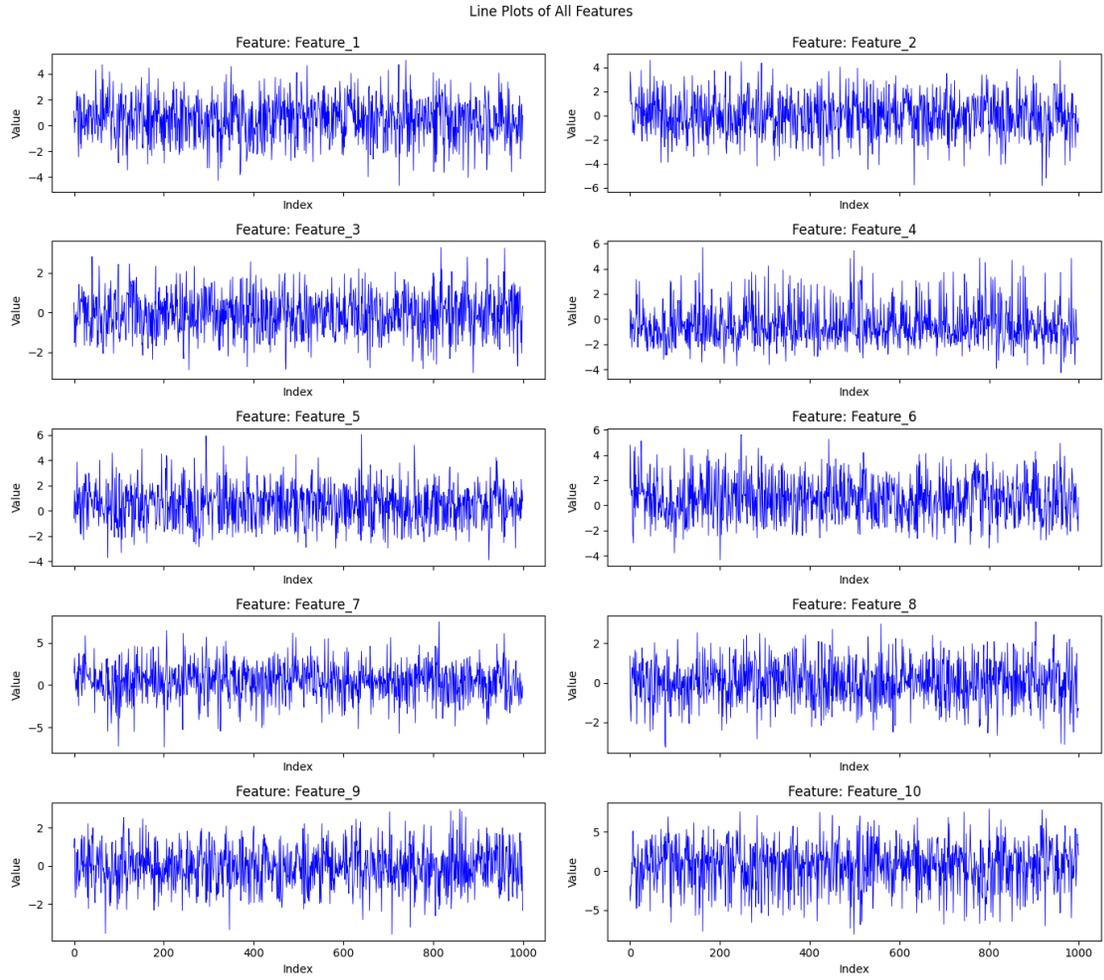

Figure 4. Histograms for each feature of the simulated dataset

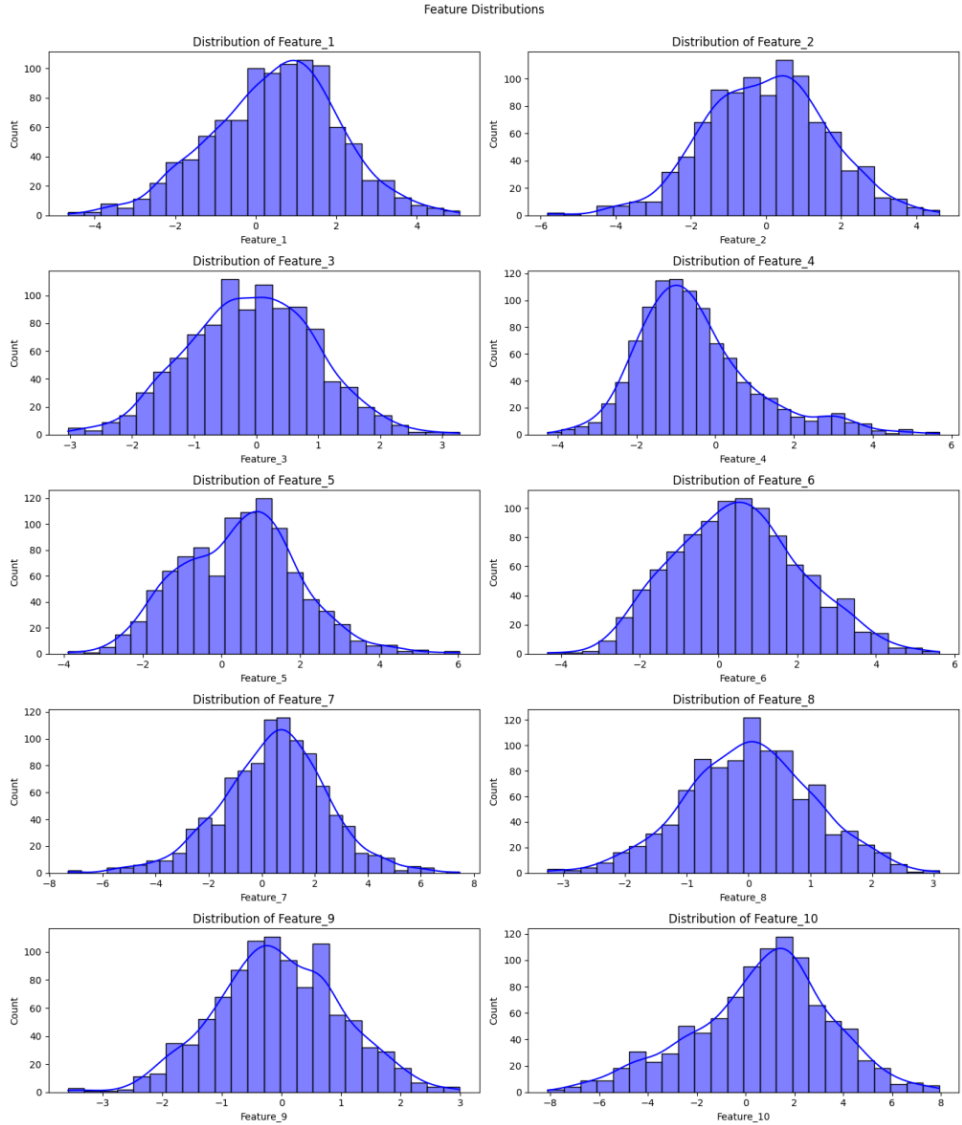

Figure 5. Histograms for each feature of the simulated dataset

## 3.1 Traditional Feature Importance

Feature Importance is a common approach to measure how the input feature affects the output. It has been used multiple times to explain faults. Permutation importance is a model-agnostic technique that assesses feature importance by shuffling (permuting) the values of a single feature and observing the impact on the model's performance (e.g., accuracy, F1 score). If shuffling a feature leads to a significant drop in performance, that feature is considered necessary. This is achieved through computing the original prediction error of the model. Then for each feature, permute the values of feature in the test dataset are permuted. Finally, the new prediction error of the model on this perturbed dataset is calculated. The importance of feature $j$ is the increase in the prediction error as a result of the permutation,

$$\text{Importance}(j) = \text{Error}_{\text{permuted}}(j) - \text{Error}_{\text{original}}$$

Aldrich and Auret [11] used permutation importance to explain the importance of input variables for fault detection and diagnosis of steady state faults.

Tree-based feature importance is specific model-specific to decision tree-based algorithms, such as Random Forests and XGBoost. It measures the contribution of each feature to the reduction in impurity (e.g., Gini impurity) or error at each node of the tree. First the Gini impurity is calculated at node $t$ by $G(t) = 1 - \sum_{i=1}^{k} p_i^2$ where $p_i$ is the proportion of the samples that belong to class $i$ at node $t$.

$$\text{Importance}(j) = \frac{\sum_{t \in T: j \text{ splits } t}(p(t) \times \Delta G(t,j))}{\sum_{t \in T} p(t) \times \Delta G(t)} \qquad (1)$$

where $p(t)$ is the proportion of samples reaching node $t$, $\Delta G(t,j)$ is the decrease in Gini impurity from splitting node $t$ on feature $j$, and $T$ is the set of all nodes that split on feature $j$ [12]. Using the simulated dataset, we apply Random Forest as a classifications tool and plot the feature importance in Figure 6. As it can be seen the feature importance plot was able to distinguish the important from non-important features.

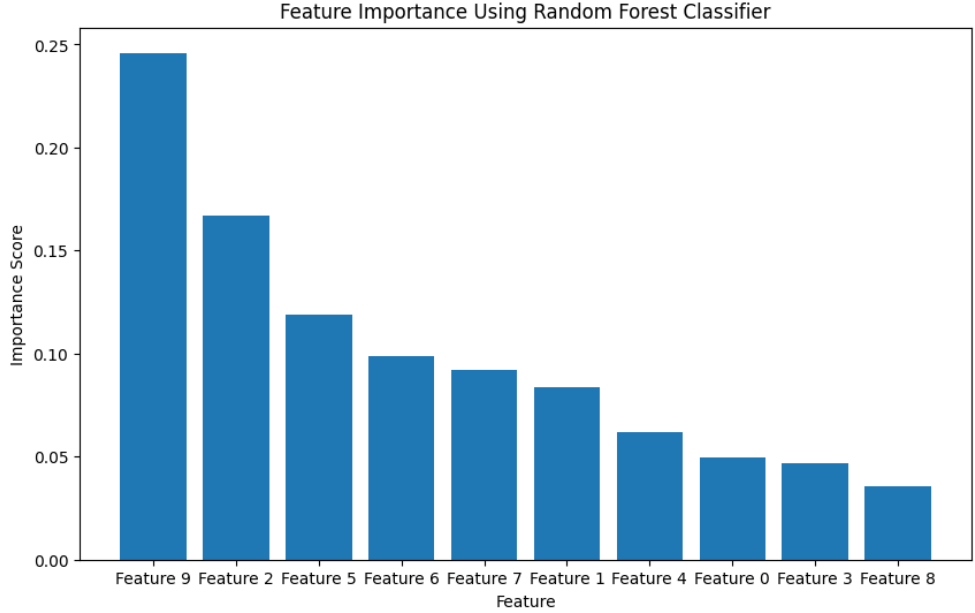

Figure 6. Feature importance using Random Forest classifier

In the context of fault detection and diagnosis, examples of explaining XGBoost, Random Forest, and Decision trees using Tree-based feature importance includes [13-19].

**3.2 Local Interpretable Model-Agnostic Explanations (LIME)**

LIME is a model agnostic method that provides explanations for individual predictions by approximating the model locally with a simpler interpretable model.

Since LIME does not use a specific model (i.e., any model can be used as an interpretable model), it does not have specific equations. For sake of clarification and assuming a linear explanation model the algorithm should work as follows [20]:

1. Select an instance for explanation which is linear as assumed,

$$f(z') = \beta_0 + \sum_{j=1}^{n} \beta_j z'_j \qquad (2)$$

where $\beta_j$ are the coefficients explaining the contribution of each feature, and $z'_j$ are the features of the perturbed sample.

2. Generate a new dataset consisting of perturbed samples around the selected instance. This is typically done by adding noise to the features of the instance.

3. Assign weights $\pi(x)$ to these new samples based on their proximity to the original instance. The closer a perturbed sample is to the original instance, the higher its weight. The weights $\pi(x)$ for the samples are computed using a kernel function that decreases with the distance from the original instance $\pi(x)$:

$$\pi(x) = \exp\left(-\frac{d(x,z')^2}{\sigma^2}\right) \quad (3)$$

where $d(x, z')$ is the distance between the original instance and the perturbed instance, and $\sigma$ is a bandwidth parameter.

4. Use the coefficients of the simple model to explain the contribution of each feature to the prediction of the instance.

Using our simulated dataset, we generate the LIME values for one instance in Figure 7 to show how explainability works in that case. The values in orange or blue reflect that this feature is affecting either the positive or negative class prediction, respectively. Features highlighted in orange contribute towards predicting the instance as belonging to the faulty class, whereas features in blue push the prediction towards the 'non-faulty' class. The values here represent the contribution of each feature to the prediction. Hence higher values mean higher effect.

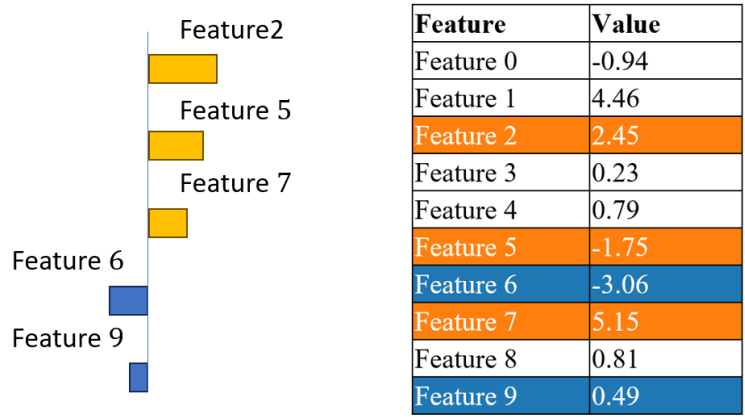

Figure 7. LIME values for explaining the simulated dataset

It is observed that the feature importance rankings derived from Random Forest and LIME showed notable differences. Random Forest offers a global perspective on feature relevance. This method emphasizes features that consistently improve model accuracy across the entire dataset. Conversely, LIME provides local explanations. This local scope can reveal the significance of features in particular contexts or for specific data points, which may not align with the broader trends captured by Random Forest.

Several studies have leveraged LIME for fault detection applications. For instance, Srinivasan, et al. [21] applied LIME to explain a model for fault detection in chillers. Lu, et al. [22] used a 1D convolutional neural network (CNN) classifier for rolling bearing health state identification, then employed LIME to interpret the model's predictions. In Sairam, et al. [23], LIME was combined with XGBoost to build an explainable fault detection model for photovoltaic panels implementable on edge nodes with a real application on photovoltaic panels. Sundarrajan and Rajendran [24] also utilized LIME to elucidate the results from pretrained models such as VGG, Inception and Resnet. Zhang, et al. [25] developed an attention-based interpretable prototypical network for small-sample damage identification with ultrasonic guided waves. Its channel attention module mitigated overfitting while extracting discriminative features. LIME then explained the network's intrinsic mechanism for damage identification by determining critical input features contributing to its predictions.

**3.3 Shapley Additive Explanations (SHAP)**

SHAP, in contrast to LIME, assigns each input feature a specific importance value based on its exact contribution to the model's predictions using concepts from cooperative game theory, namely Shapley values.

The Shapley value is a method that assigns a value to each feature that reflects their contribution to the difference between the prediction for a particular instance and the average prediction over the dataset. Shapley value can be represented as [26]:

$$\phi_i(f, x) = \sum_{S \subseteq N \setminus \{i\}} \frac{|S|! \, (|N| - |S| - 1)!}{|N|!} \left( f_x(S \cup \{i\}) - f_x(S) \right) \qquad (4)$$

Where $N$ is the set of all features, $S$ is a subset of features excluding feature $i$, $|S|$ is the number of features in subset $S$, $f_x(S)$ is the prediction of the model using only the features in set $S$, and $f_x(S \cup \{i\})$ is the prediction of the model using the features in $S$ plus feature $i$. The factorial terms $\frac{|S|!(|N|-|S|-1)!}{|N|!}$ represent the number of possible sequences that can form $S$ without including feature $i$.

SHAP is built upon Shapley where $\text{explanation}(x) = \phi_0(f) + \sum_{i=1}^{M} \phi_i(f, x)$ where: $\phi_0(f)$ is the average prediction for the dataset and $\sum_{i=1}^{M} \phi_i(f, x)$ sums the SHAP values for all features, quantifying their individual contributions to the difference between the model's prediction for $x$ and the average prediction. Thus, for a simple linear regression model, the SHAP values for each feature in a particular prediction instance would directly correspond to weights multiplied by the deviation of the input from its mean, reflecting each feature's contribution towards the deviation from the average prediction.

Reflecting to our simulated dataset, we use SHAP and generate explanations as presented in Figure 8. Features with a wide spread of SHAP values, such as Feature 9, have variable impacts on the model's predictions (i.e., significantly affect the output), depending on their values in specific instances. Notably, SHAP shows similar results as those from Tree-based feature importance.

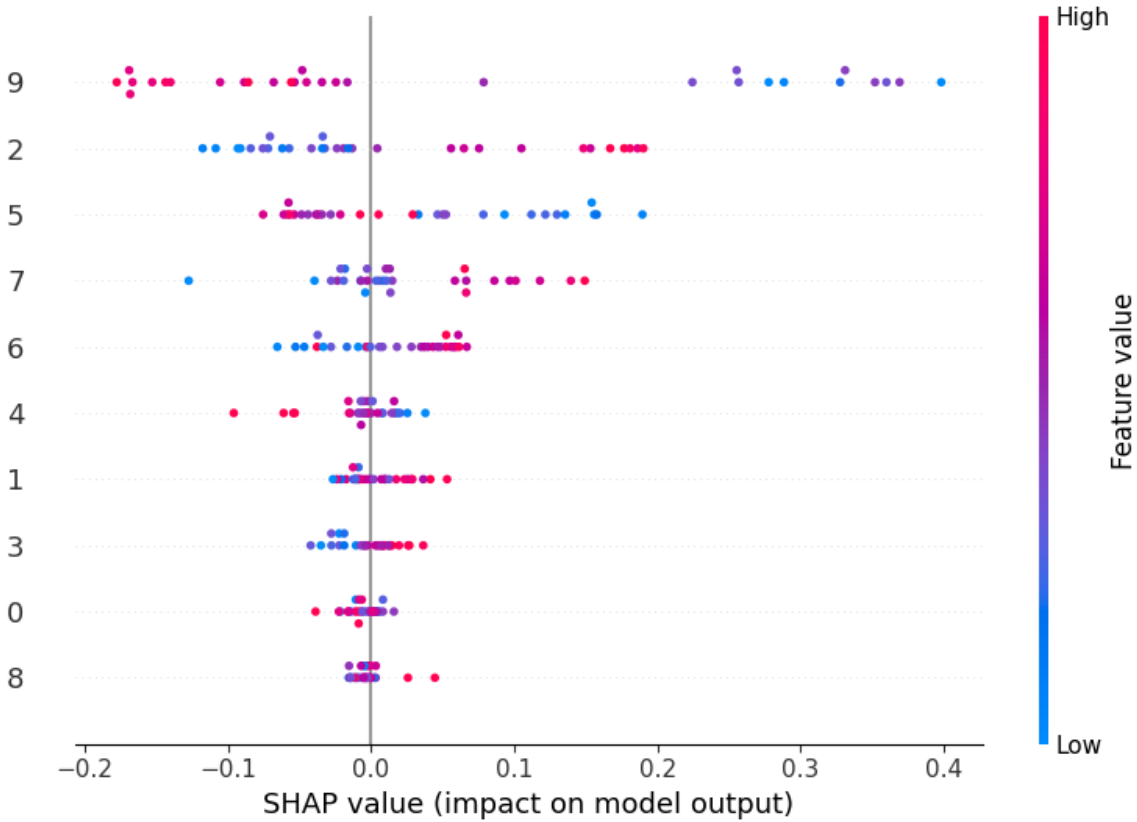

Figure 8. SHAP values for explaining the simulated dataset

The efficacy and potential of SHAP have been highlighted by numerous studies that have utilized its interpretations in fault detection and diagnosis applications. This approach not only enhances the transparency of AI models but also supports more informed decision-making by highlighting the significant factors influencing model predictions.

Hwang and Lee [27] employed SHAP to analyze and interpret feature contributions in a bidirectional LSTM (Bi-LSTM) anomaly detection model for industrial control systems, supporting timely operational responses. Jang, et al. [28] utilized SHAP values with an adversarial autoencoder to build fault maps distinguishing various fault types in chemical processes. The method involves creating an adversarial auto-encoder model and monitoring an index that is based on a weighted average of both the hoteling and squared prediction errors. Then, SHAP values are calculated and analyzed. Accordingly, a fault map is introduced. The proposed method is applied to two chemical process systems. The results demonstrated that the proposed method accurately diagnoses single and multiple faults and can distinguish the global pattern of various fault types. Chowdhury, et al. [29] used SHAP to diagnose faults in 3-D printers after using ensemble learning model of Random Forest and XGBoost for the training.

Brusa, et al. [30] showcased the effectiveness of SHAP in discerning and elucidating the most pivotal features in the context of two models: the Support Vector Machine (SVM) and the k-Nearest Neighbor (kNN). These models were employed to analyze vibration data originating from medium-sized bearings, which were collected and scrutinized for the purpose of classifying rotor faults. Moreover, Choi, et al. [31] incorporated SHAP explanations within an online framework for predictive maintenance in chemical plants.

Baptista, et al. [32] utilized the SHAP model to analyze the outcomes of three algorithms: Linear Regression, Multi-Layer Perceptron, and Echo State Network, with the goal of determining if there is a correlation between the prognostics metrics and the SHAP model's explanations. They used a baseline dataset containing jet engine run-to-failure trajectories. The results demonstrated a close tracking between SHAP values and the metrics, with differences observed among the models of course. Zhang, et al. [33] proposed a gamification approach named ENIGMA to address the security challenges in Cyber-Physical Systems. This approach uses Digital Twins as both a security assessment tool and a training platform, where game scenarios involve human and AI participants representing attackers and defenders, respectively. Then SHAP values are used to clarify AI decisions concerning attack vectors. Similarly, Kumar and Hati [34] employed SHAP methodology to elucidate induction motor fault classification by CNNs

Onchis and Gillich [35] combined LIME and SHAP techniques to generate an explainable prediction and enhance the interpretability of deep learning models used for classifying accelerometer data. The authors introduced a compound stability-fit compensation index to address the instability of local explanations and ensure the quality of the explanations by incorporating LIME and SHAP, the study aimed to accurately characterize the location and depth of damaged beams in a transparent and trustworthy manner.

Brito, et al. [36] utilized SHAP for fault diagnosis in rotating machinery. Furthermore, the authors conducted a comparison between SHAP and Local Depth-based Feature Importance for the Isolation Forest (Local-DIFFI) in terms of the effectiveness of these models in providing insights and explanations for fault diagnosis in rotating machinery. The results revealed that SHAP was a valuable tool for fault diagnosis. Yet, the authors did not reach a decisive conclusion on which is best and suggested the choice of the explainability model to be based on a trade-off between response time and precision. The authors showcased the effectiveness of SHAP in diagnosing faults on three datasets containing different mechanical faults in rotating machinery.

## 3.4 Partial Dependence Plot (PDP) and Individual Conditional Expectation (ICE)

Partial Dependence Plots (PDP) are model-agnostic method designed to shed light on how individual features influence predictions. They operate by graphically illustrating the average effect of changing a single input feature while holding all other features constant, enabling a global view of feature importance and its relationship with model predictions. Distinguishably from LIME, SHAP, and Permutation Feature Importance, PDPs offer a broader perspective, focusing on the overall behavior of the feature throughout the model's prediction space, making them suitable for assessing feature impact across various data points and providing insights into trends and patterns.

In contrast, ICE generates plots for individual data points, visualizing how predictions shift as a feature alters while fixing other inputs. In essence, ICE offers an individualized view of the feature's effect on model predictions.

While PDP captures general behavior, ICE furnishes a localized understanding by showing per-example impacts. This is particularly useful for detecting variations and heterogeneity in how the feature influences different data points, revealing distinctions that might not be apparent. ICE plots are beneficial when it is needed to examine how a specific instance or subgroup is impacted by a particular feature, making it a valuable tool for fine-grained model interpretation and understanding the local behavior of the model.

The math behind PDP and ICE is straight forward. For a feature $X_s$ and a set of features $X_c$ (complementary set), the partial dependence function for $X_s$ is defined as:

$$\text{PD}_s(x_s) = E_{X_c}[f(x_s, X_c)] = \int f(x_s, x_c)\, p(x_c)\, dx_c \qquad (5)$$

where $f(x_s, X_c)$ represents the model prediction function, $x_s$ is a specific value for the feature $X_s$, $X_c$ are the other features in the dataset. $p(x_c)$ is the probability density function of $X_c$. The expectation $E_{X_c}$ is calculated over the distribution of the other features, effectively marginalizing out these features. Intuitively speaking, this means we calculate the expected value of the model predictions over the distribution of the other features, keeping $x_s$ fixed at $x_s$.

For each individual $i$ in the dataset, the ICE plot for feature $X_s$ is defined as:

$$\text{ICE}_s^i(x_s) = f(x_s, x_c^i)$$

where $f(x_s, x_c^i)$ is the prediction of the model when $X_s$ is set to $x_s$ and all other features $X_c$ are fixed at their actual values $x_c^i$.

Using the simulated dataset, we applied PDP and ICE to generate explanations, shown in Figure 9. Each subplot corresponds to a different feature. PDP is the average effect of a feature on the prediction outcome across the dataset. The more nonlinear the line, the higher influence it has on the class predictions. On the other hand, ICEs are the blue lines plotted for each instance in the dataset showing how the prediction changes with different values of the feature while other features are held constant. This provides a more granular view than PDP, highlighting individual variations and potential anomalies. In our simulated datasets, features like 9, 7, 6, 5, and 2 display a wide range of PDP and ICE values, have varying impacts on the model's predictions based on their specific instance values.

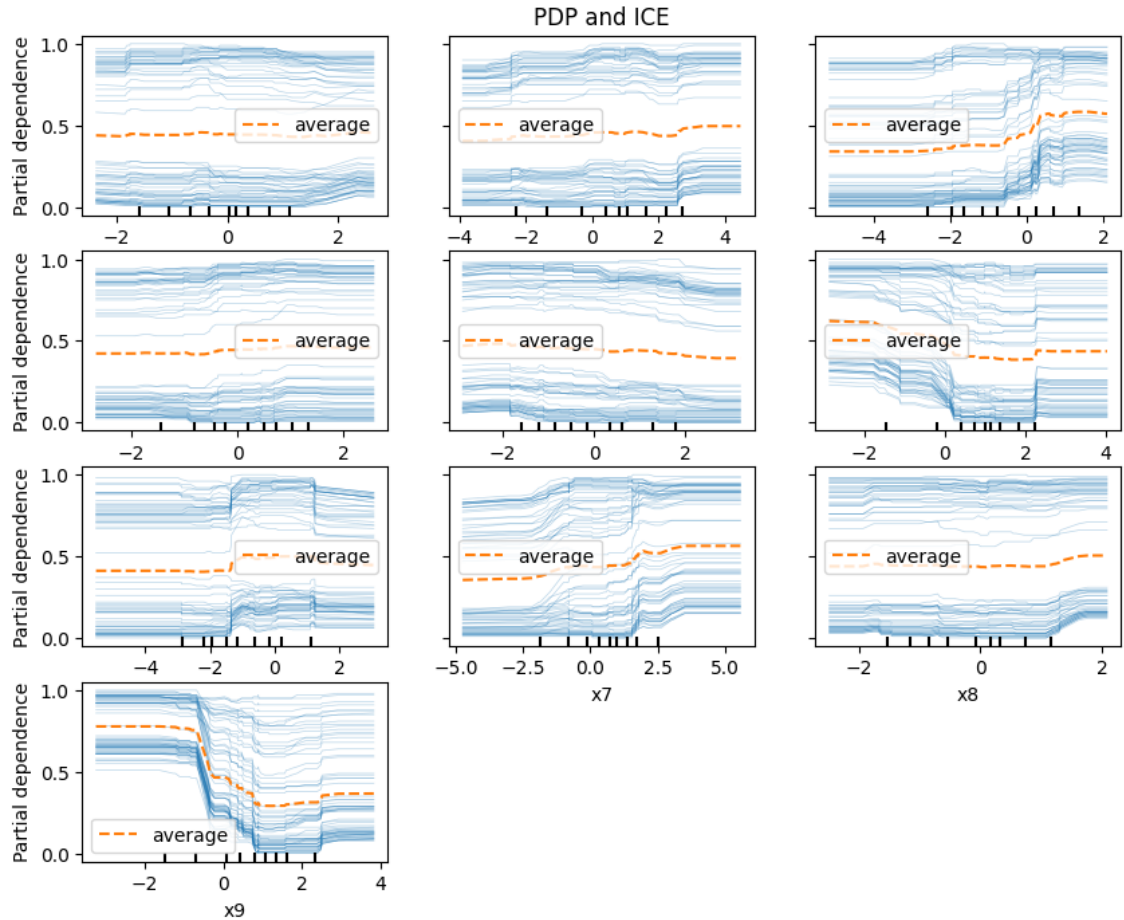

Figure 9. PDP and ICE curves for the simulated dataset

Several studies leverage these techniques to better comprehend fault detection solutions, as shown. Prakash, et al. [37] proposed a methodology for detecting internal leakage in hydraulic pumps using an unbalanced dataset of electrical power signals from the pump's drive motor. The methodology involves extracting refined composite multiscale dispersion and fuzzy entropies, along with three statistical indicators and second-order polynomial-based features. PDP and ICE are used to visualize and normalize these features.

In Mehdiyev and Fettke [38], ICE and SHAP were utilized to help in predictive process monitoring. By integrating top-floor and shop-floor data and applying a deep learning model, process outcomes were predicted. The authors showed that generated explanations using ICE and SHAP allowed domain experts to examine different perspectives and understand the factors influencing the predictions.

Danesh, et al. [39] exploited PDP and ICE to visualize and explain the impact of inputs on predictions from a neural network used to predict electrical power output in a combined cycle power plant. Kang, et al. [40] used a decision tree machine learning model for extracting failure features of automatic train protection systems. The model uses parameters such as system type, operation mileage, and service time to predict the cumulative failure rate. For interpretability, the authors used PDP, ICE, Feature Importance, and SHAP to reveal the logic behind the model learning as well as the importance of each of the features that affect the failure rate.

**3.5 Layer-wise Relevance Propagation (LRP)**

LRP is a technique that offers valuable insights into the inner workings of neural networks, shedding light on the factors influencing their predictions. LRP achieves this by unraveling the model's predictions and attributing them to individual input features. Its fundamental premise revolves around the dissecting the model's output and subsequently allocating relevance scores to neurons and input features across the network's layers.

One of the primary goals of LRP is to facilitate a deeper comprehension of the neural network's decision-making process, specifically by unveiling the significant roles played by different input features. LRP was widely used in different applications. For instance, Agarwal, et al. [41] introduced a deep learning-based statistical monitoring methodology for Fault Detection and Diagnosis (FDD). It was found that the relevance scores, generated through LRP, could be employed iteratively to identify, and eliminate redundant input feature vectors or variables. The consequence of this curation was a reduction in over-fitting on noisy data, an enhancement in the distinction between output classes, and ultimately, superior FDD test accuracy. In another study conducted by Grezmak, et al. [42], LRP was applied to scrutinize the performance of a CNN used for machine fault diagnosis, specifically through the analysis of time-frequency spectra images derived from vibration signals. LRP's role here was to provide a pixel-level representation of the input signal, revealing which values held the most sway over the diagnosis outcomes.

In a manufacturing context, Lee, et al. [43] utilized LRP to explore the predictive capacity of a defect image classification model. This allowed for the visualization and highlighting of the relevant regions within defect images, offering domain experts valuable insights into the model's decision-making process. Similarly, Han, et al. [44] employed LRP to explore the impact of data on a trained deep learning model for mechanical motor fault diagnosis.

### 3.6 Class Activation Mapping (CAM)

CAM, is a technique, that empowers (CNNs) to provide interpretability by generating heatmaps, thus shedding light on the specific regions within an input image that significantly influences the model's predictions. The hallmark of CAM is its ability to utilize global average pooling to weigh feature maps, thereby granting an understanding of the areas in an image that hold the utmost sway over the model's decision.

Chen and Lee [45] embarked on a study focused on fault detection using vibration signals, which are converted into images through Short-Time Fourier Transform (STFT). Employing a CNN for classification, they utilized the Gradient-weighted CAM (Grad-CAM) technique to elucidate classification decisions. Grad-CAM, a variant of CAM distinguished by its gradient-based approach, calculates the gradients of the predicted class with respect to feature maps in a convolutional layer, resulting in a heatmap that highlights the image regions most influential to the model's output. In this study, attention was conducted with neural networks, Adaptive

Network-based Fuzzy Inference Systems (ANFIS), and decision trees to underscore the proposed results. Similarly, Kim and Kim [46] integrated Grad-CAM into a CNN-based bearing fault diagnosis technique. They used the Normalized Bearing Characteristic Component (NBCC) as the CNN input, representing bearing failure symptoms effectively. Yoo and Jeong [47] also presented a vibration analysis process for bearing fault diagnosis using a fine-tuned VGG-19 model and Grad-CAM. The vibration data were collected from a motor using an accelerometer and Internet of Things (IoT) module. Spectrogram images were then generated for normal and different fault situations. The visualization model they proposed was compared to conventional defect frequency analysis methods.

Alike, Brito, et al. [48] applied Grad-CAM with a 1D CNN to enhance explainability in the context of rotating machinery. Yu, et al. [49] used Grad-CAM and Eigenvector-based Class Activation Map (Eigen-CAM) to interpret ResNet06 in various databases, including bearing and gearbox datasets.

Kim, et al. [50] introduced an explainable model for fault diagnosis in linear motion guides using time-domain signals. The Frequency-domain-based Grad-CAM (FG-CAM) method was utilized to visualize the model's classification criteria learned in the time domain. The authors highlighted the challenge of applying the model to multiaxis data, emphasizing the need to extract features from each axis individually. This led to the development of a grouped convolution approach, enabling the extraction of feature maps from each X-, Y-, and Z-axis and facilitating the visualization of the importance of different frequencies on model decisions. Furthermore, multivariable data-based FG-CAM (mFG-CAM) was proposed to visualize decision criteria with multiaxis vibration signals.

Li, et al. [51] introduced Multilayer Grad-CAM (MLG-CAM) to tackle the issue of decreasing feature resolution in Grad-CAM. The MLG-CAM leverages gradients across multiple convolutional layers to obtain activation maps with varying resolutions, which are then combined through layer-weighted summation to generate a comprehensive activation map. Experiments conducted with MLG-CAM demonstrated its ability to emphasize cyclo-stationary impulses in the time domain and fault characteristic frequencies in the frequency domain.

Oh and Jeong [52] proposed a framework incorporating CNN for fault detection and CAM for fault diagnosis. To enhance the reliability of the diagnostic results, process monitoring was performed using Variational Autoencoders (VAE), which learned the CAM produced in the fault detection and diagnosis process, treating CAMs generated with misclassified label information as anomalies.

Lee, et al. [53] introduced a novel data imagification approach called Fuzzy-based Energy Pattern Image (FEPI) generation, which transformed sensor signals into FEPI data. A CNN-based fault diagnostic model was trained using FEPI data, and Grad-CAM was used to interpret the model's predictions and identify critical regions for fault classification. This approach was presented in a case study of robotic spot welding, yielding promising results.

Yang, et al. [54] proposed a different fault detection and diagnosis method in rotating machinery. The main structure of the proposed method is based on the standard CNNs, but they added a penalty term to the loss function to penalize the model if it learned some insignificant fault features in the training process. They further utilized smoothed score-CAM, an upgraded version of CAM, to improve localization accuracy and heatmap visual quality. The smoothing process involved applying a Gaussian filter to the activation map to reduce over-activation and noise, resulting in more refined and reliable heatmaps.

Gwak, et al. [55] employed a frequency-domain-based architecture for fault classification using preprocessed vibration signals. Their method included FG-CAM for visualizing decision criteria in the frequency domain, allowing for the identification of critical input frequency components. Additionally, they utilized the Power-Perturbation-Based Decision Boundary Analysis (POBA) framework to analyze changes in decision boundaries by perturbing power spectral densities in input vibration signals. Finally, an ensemble model was created to increase robustness based on decision boundary information obtained from POBA.

**3.7 Case Based Reasoning (CBR)**

CBR explains the decisions made by machine learning models by retrieving and adapting similar cases from a case base. In other words, it compares new instances to previously seen examples to provide explanations based on similarity to known cases. For example, suppose a machine learning model classifies a machine as faulty, in that case, CBR can be used to retrieve similar points or instances that have been previously classified as defective and provide an explanation based on the similarities between the two samples or instances. The explanation might highlight particular features or patterns in the data that led to this specific classification, helping to make the decision more transparent.

The mathematical formalization of CBR primarily revolves around defining and computing the similarity between cases. CBR itself is less about complex mathematical formulas and more about a methodology or strategy for problem-solving that involves matching new problems to previously encountered cases to find solutions. However, the crucial part that can be mathematically defined is the similarity measure, which is fundamental for retrieving the most relevant cases. One could summarize the steps for CBR as follows:

1. Represent each case in CBR as a vector in a feature space. If $x$ is a new problem and $c_i$ represents a case in the case base, both $x$ and $c_i$ can be expressed as vectors:

$$x = (x_1, x_2, \ldots, x_n), \quad c_i = (c_{i1}, c_{i2}, \ldots, c_{in})$$

2. The similarity measure $Sim(x, c_i)$ quantifies how close or similar a stored case $c_i$ is to the new problem $x$. There are various ways to define this similarity, with the choice often depending on the specific application. A common choice is to use the inverse of the Euclidean distance or Cosine similarity giver respectively as

$$Sim(x, c_i) = \frac{1}{\sqrt{\sum_{j=1}^{n}(x_j - c_{ij})^2}}, \qquad (6)$$

$$Sim(x, c_i) = \frac{x \cdot c_i}{|x||c_i|} = \frac{\sum_{j=1}^{n} x_j c_{ij}}{\sqrt{\sum_{j=1}^{n} x_j^2}\sqrt{\sum_{j=1}^{n} c_{ij}^2}} \qquad (7).$$

3. The retrieval process can be mathematically defined as selecting the case $c_i$ that maximizes the similarity measure:

$$c^* = \arg\max_{c_i \in C} Sim(x, c_i)$$

For our hypothetical simulated dataset, CBR would mean that when a new fault is detected, the system analyzes the current sensor readings, compares them to historical data, identifies the most similar previous incidents, and suggests a proven solution based on past outcomes.

Many authors have employed CBR. For example, Khosravani, et al. [56] used it in fault detection in injection molding of drippers. Zhao, et al. [57] utilized the case-based reasoning method for fault detection and diagnosis of the Tennessee Eastman process. The authors also proposed a case maintenance strategy to avoid redundant and noisy cases that are added to the case base. Similarly Boral, et al. [58] used CBR in a framework for fault detection and diagnosis and for suggested maintenance actions. An interesting study by Chen, et al. [59] proposed a CBR system based on 143 cases with the knowledge of correctly diagnosed and successfully resolved aero-engine faults. They also proposed a case similarity measure for fault diagnosis based on three local similarity measures associated with different attributes.

**3.8 Alternative and Supplementary XAI Techniques for Fault Detection and Diagnosis**

While this paper has primarily focused on utilizing XAI techniques for accurate fault detection and diagnosis, it is important to acknowledge the existence of various alternative and supplementary methods that have been applied in the field. These methods, although less mainstream or rigorously structured, have been used to address specific challenges or gain insights in fault detection and diagnosis.

In some instances, researchers have resorted to what can be termed "ad-hoc" techniques. These ad-hoc methods are not characterized by a standardized framework or a widely recognized approach but are instead devised on a case-by-case basis to cater to specific needs or unique challenges in fault detection. They may involve creative combinations of algorithms, custom feature engineering, or domain-specific heuristics. While ad-hoc techniques may lack the formal structure and interpretability associated with XAI, they have played a valuable role in addressing nuanced or domain-specific problems.

Furthermore, a range of other supplementary techniques have been applied in fault detection and diagnosis. These approaches may include traditional statistical methods, signal processing techniques, or domain-specific expert systems. These supplementary techniques, although not classified as XAI, provide important context and historical relevance in the field. Researchers have employed them to enhance the performance of fault detection systems, verify XAI-derived insights, or address challenges where interpretable models may not suffice.

Hence, in this section, we will explore these alternative and supplementary techniques, shedding light on their applications, strengths, and limitations in the broader landscape of fault detection and diagnosis.

Utama, et al. [60] used a model-agnostic method called Anchors to build an explainable fault detection model for photovoltaic panels implementable on edge nodes with a real application on photovoltaic panels. The purpose of Anchors is to find a decision rule that approximates the decision function of the model around that individual data point. These anchors are more human-interpretable explanations that can be mapped back to data features. Anchors can thus identify the key patterns in sensor data that are indicative of a fault condition according to the model.

Further advancing the field, Kim, et al. [61] proposed a visual XAI method that aims to provide explainability in fault diagnosis using a 1D vibration signal. It introduced a Frequency Activation Map (FAM) to visualize the classification criteria of a 1D CNN model. The methodology involves designing a CNN model with a norm constraint on the filters to ensure consistent filtering of frequency information. The model learns from vibration signals to classify normal and faulty states of equipment. The FAM is generated to visualize the specific frequencies that the model focuses on for classification. It highlights characteristic frequencies associated with normal and faulty states, providing insights into the model's decision-making process. In a similar vein, Kolappan Geetha and Sim [62] investigated misclassifications of a CNN model used for crack classification. They mapped these misclassified instances back to the t-Distributed Stochastic Neighbor Embedding (t-SNE) metadata space to analyze how crack signatures are clustered with non-crack and vice versa, leading to misclassifications. t-SNE is a technique used to visualize high-dimensional data by reducing its dimensionality while preserving local relationships. It maps the data to a lower-dimensional space, making it easier to understand and interpret complex patterns and relationships in the data. By examining the spatial distribution of the features in the t-SNE visualization, they gain insights into the reasons behind misclassifications.

Building upon the theme of interpretability, Jiang, et al. [63] proposed an interpretable DL model named Multi-Wavelet Kernel Convolution Neural Network (MWKCNN) for fault diagnosis. First, a feature extraction layer named the Multi-Wavelet Kernel Convolution (MWKC) layer was constructed based on the continuous wavelet transform (CWT) to locate and detect the impulse signatures of faults. It can capture sufficient feature representations by involving signal processing knowledge. Then, a Kernel Wights Recalibration (KWR) module was used to assign different weights to different wavelet kernels dynamically. The interpretability of the proposed method was achieved by analyzing the peculiarity of different wavelet kernels, the variation of kernel weights, and the visualization of learned features. The performance of the proposed method was validated using two gearbox datasets.

Transitioning to XAI in fault detection using process monitoring, Lu and Yan [64] proposed a Deep Fisher Auto Encoder based Self-Organizing Map (DFAE-SOM) model for visual process monitoring. The method combines Fisher Discriminant Analysis (FDA) and auto encoders to create Fisher Auto Encoder (FAE), which helps extract discriminative features and reduces reconstruction error. By stacking the FAE models, a Deep Fisher Auto Encoder (DFAE) architecture was created. To make DFAE more interpretable, it was combined with a self-organizing map to project the high-dimensional data into a 2D space, where normal and fault classes are visualized in separate regions.

In the context of explainable rule based decision-making, Dorgo, et al. [65] proposed a decision tree-based classifier for fault classification. The model utilized a sliding window-based data preprocessing approach and was designed to be able to detect and isolate faulty states. The proposed method in the paper used alarm thresholds to trigger alarms when the process variables exceed certain limits. The decision tree classifier was then employed to analyze the situation further and provide a set of rules and conditions that can be followed to interpret the alarm and make decisions. Similarly, Obregon, et al. [66] proposed a rule-based explanations (RBE) framework in combination with machine learning interpretation methods to understand the decision mechanisms of accurate and complex predictive models, specifically tree ensemble models, in the context of plastic injection molding quality control. The framework generated simple decision rules along with partial dependence plots and feature importance rankings to provide meaningful explanations and enhance the understanding of the main factors influencing manufacturing quality. The applicability of the RBE framework was demonstrated through two experiments using real industrial data from a plastic injection molding machine, showcasing its potential for improving production efficiency in this domain.

Harinarayan and Shalinie [67] took a different tack by leveraging SHAP for both local and global explanations in an XGBoost model. Then they generated diverse counterfactual explanations to be used as action recommendations to correct the fault scenario. Counterfactual explanations are a type of XAI technique. They involve generating alternative scenarios that could have led to a different outcome, given the same input data. It works by optimizing a distance function that provides the minimum change to the input instance to get a different target output. The decision maker can then choose the optimal change required in terms of cost.

In the field of probabilistic models, Maged and Xie [68] presented a systematic approach for utilizing the prediction uncertainty information generated by Bayesian Neural Networks (BNN) models along with the prediction values obtained from the output layer of the network in order to make optimal decisions. BNNs can provide some interpretability through their probabilistic nature, which allows for uncertainty quantification. However, they are not inherently interpretable models like decision trees or rule-based systems. Their interpretability is limited compared to more explicitly interpretable models in the field of XAI. The proposed approach is applied to a real case study on vertical continuous plating of printed circuit boards.

Lastly, Conde, et al. [69] incorporated interpretability constraints into a boosting algorithm to produce accurate and easily interpretable classification rules. The authors introduced two methods for binary classification. These methods, Simple Isotonic LogitBoost (SILB) and Multiple Isotonic LogitBoost (MILB), aim to create classification rules that align with known monotonic relationships within the data. SILB operates by selecting the best-fitting variable in each boosting step while carefully considering isotonicity, ensuring that the relationships between selected variables and classification outcomes follow a consistent trend. MILB, on the other hand, takes a different approach, refitting the entire problem in each boosting step. This means that all predictors change their roles in the classification rule during the process while still respecting isotonicity constraints. Their isotonic boosting approach was evaluated using simulations and real-world induction motor failure data.

## 3.9 Choosing the Right XAI Technique

Striking an optimal balance between model accuracy and explainability represents a significant challenge in machine learning, particularly in the context of fault detection systems. The capability to predict faults in unseen data, as measured by model accuracy, frequently conflicts with the requirement for model explainability, which involves understanding the rationale behind specific fault predictions. Although achieving this balance is difficult, various strategies exist that can improve the explainability of fault detection models without substantially affecting their performance. These strategies provide valuable insights into the decision-making processes of AI systems, potentially enhancing their usability and trustworthiness in critical applications.

In terms of accuracy, complex machine learning architectures, such as deep neural networks, demonstrate excellence in discerning intricate fault patterns but often function as inscrutable black-boxes. For instance, a deep neural network might achieve state-of-the-art accuracy in fault detection but struggle to provide meaningful insights into its decision-making process, making it challenging for operators to understand the root cause of detected faults.

Conversely, from an explainability perspective, simpler models such as decision trees and their linear counterparts provide high levels of interpretability, though this may come at the expense of reduced accuracy in fault detection. In scenarios where fault detection in machinery is critical, a decision tree can offer clear and transparent criteria for identifying faults. However, its predictive accuracy may not match that of more complex models. This trade-off highlights the ongoing challenge in machine learning of balancing the accessibility of the model's reasoning with its effectiveness in performing critical tasks.

The trade-off between accuracy and explainability in fault detection systems is contingent upon the nuances of the task. Interpretability assumes ascendancy in domains like industrial machinery, where understanding the rationale behind detected faults is crucial for maintenance personnel. In contrast, in fault detection systems for critical applications like aerospace, accuracy might be prioritized to ensure timely and precise identification of potential issues, allowing for prompt corrective actions.

Following, we provide guidance to users seeking to incorporate XAI techniques into their analysis. Table 1 offers an overview of general questions and engineering-specific questions related to machine behavior and potential methodologies for interpretation. Each methodology is associated with suitable implementation processes, including relevant libraries in both R and Python.

To illustrate the practical application of these concepts, consider a scenario on a production floor where "machine id 1" begins displaying erratic behavior. This sudden change raises concerns about the potential failure of a specific component. To diagnose the issue and facilitate informed decision-making, stakeholders' resort to XAI techniques.

Table 1. Guide of the selection of the suitable XAI technique

| General Question | Engineering Question | Methodology to be Used | Implementation Process |
|---|---|---|---|
| What is the Influence of features on model output? | How do various features influence the model's output? | SHAP | SHAP (Python library), SHAPR (R Package), IML (R Package), shapviz (R Package), dalex (Python and R Package) |
| How does the system perceive the impact of varying the value of a specific feature on the model output? | How does the system view the changes in feature values? | PDP / ICE | PDP (R Package), PDPbox (Python library), ICEbox (R Package) |
| How does the model draw upon past cases to justify the prediction? | Why does the system predict the failure of "machine id 1" based on past cases? | LIME | LIME (Python library, R Package), lime (Python Package), iml (R Package) |
| How does the system determine and articulate the most influential and reliable features guiding a specific prediction or decision? | What is the vibration feature threshold to ascertain correct machine functionality, and how accurate is the system in determining this threshold? | Anchors | Anchors (Python library), alibi (Python library) |

| What guiding rule or precedent does the system follow in making decisions? | What rule does the system adhere to when making decisions about the machinery? | Feature importance (random forest or XGBoost) | Scikit-learn (Python library), Xgboost (R Package), Lightgbm (R Package), ranger (R Package), h2o (Python and R Package) |
|---|---|---|---|
| How does the model output change with variations in a specific feature, considering past cases and their outcomes? | How does the model output fluctuate with variations in a specific feature considering historical cases? | Case-Based Reasoning | CBR (R Package), cbr (Python library) |
| What input features are most relevant to a prediction? | How does the system visualize and identify the most relevant component in "machine id 1" for a given prediction? | Class Activation Mapping | CAM (Python library), tf-keras-vis (Python library), keras-vis (Python library) |

In addition to the specific XAI techniques outlined above, it is essential to recognize the users' flexibility in employing alternative methods for model interpretation. Exploratory data analysis (EDA), Bayesian methods, or ad-hoc approaches can serve as valuable supplements to the suggested XAI techniques. These traditional approaches enable users to utilize statistical and probabilistic reasoning, exploring relationships within the data and providing additional context to model predictions. The selection of a particular method ultimately depends on the nuances of the analysis and the desired level of interpretability goals.

## 4. Conclusion and Expanding Horizons in XAI

As discussed above, the primary challenges when considering XAI revolve around the inherent trade-off between explainability and performance. While interpretable models offer clarity, they may attain a different the same level of accuracy than as their more complex counterparts, necessitating a delicate balance between transparency and predictive power. Another challenge arises from the subjectivity inherent in explanations. Different users may demand varied types and levels of explanations, posing a hurdle in developing universally acceptable elucidations. Balancing this diversity of needs becomes essential for the broader adoption of explainable ML.

Scalability is a pertinent limitation, with some explainable ML methods facing challenges adapting to large datasets or high-dimensional input spaces. The constraints on scalability impact the applicability of specific explanation techniques, prompting the exploration of more scalable alternatives. Finally, evaluating the quality and usefulness of explanations introduces complexity as it often relies on subjective human judgment. This intricacy underscores the need for standardized metrics and evaluation frameworks to enhance the reliability and comparability of different explainability approaches.

As we have seen in the literature, conventional approaches often confine explanations to global or local perspectives in the context of fault detection and diagnosis. However, the field is undergoing a transformative phase, introducing innovative methods to address existing challenges. Integrating explainability into models like Generalized Additive Models (GAMs) is a breakthrough, allowing the transparent modeling of non-linear relationships between faults and measures for enhanced interpretability. Additionally, the Deep Taylor Decomposition (DTD) emerges as a sophisticated method, deconstructing deep learning model outputs by attributing contributions to each neuron. This provides a clear understanding of how specific components influence overall model predictions. Another captivating development involves explaining models in natural language, encompassing methods that generate human-understandable narratives in text, speech, or images. These advancements signify a pivotal shift in explainable fault detection and diagnosis, offering solutions to limitations and broadening the understanding and adoption of transparent AI approaches. An additional significant advancement in the field of XAI is the development of transparent neural networks. These are graph-based computational models explicitly designed to enhance human comprehension. This approach very promising since it introduces an algorithm for automatic network development through environmental interactions. In other words, one can dynamically add and/or remove structures to facilitate spatial and temporal memory, culminating in the automatic generation of a comprehensive computational model. It combines concept formation with various forms of reasoning—deductive, inductive, and abductive—representing a substantial step towards creating AI systems that are both transparent and understandable.